\documentclass[letterpaper]{article} 
\usepackage[]{aaai23}  
\usepackage{times}  
\usepackage{helvet}  
\usepackage{courier}  
\usepackage[hyphens]{url}  
\usepackage{graphicx} 
\usepackage{natbib}  
\usepackage{caption} 
\usepackage{algorithm}
\usepackage{algorithmic}
\usepackage{booktabs}

\usepackage{newfloat}
\usepackage{listings}
\DeclareCaptionStyle{ruled}{labelfont=normalfont,labelsep=colon,strut=off} 
\floatstyle{ruled}
\newfloat{listing}{tb}{lst}{}
\floatname{listing}{Listing}
\usepackage{amssymb}
\usepackage{multirow}
\nocopyright 

\usepackage{marvosym}
\title{PersDet: Monocular 3D Detection in Perspective Bird's-Eye-View}
\author{
    Hongyu Zhou,
    Zheng Ge\textsuperscript{\Letter},
    Weixin Mao,
    Zeming Li
}
\affiliations{
    MEGVII Technology\\
    \{zhouhongyu,gezheng,maoweixin,lizeming\}@megvii.com
}

\begin{document}
\maketitle
\begin{figure*}[h]
	\centering
    \includegraphics[width=\textwidth]{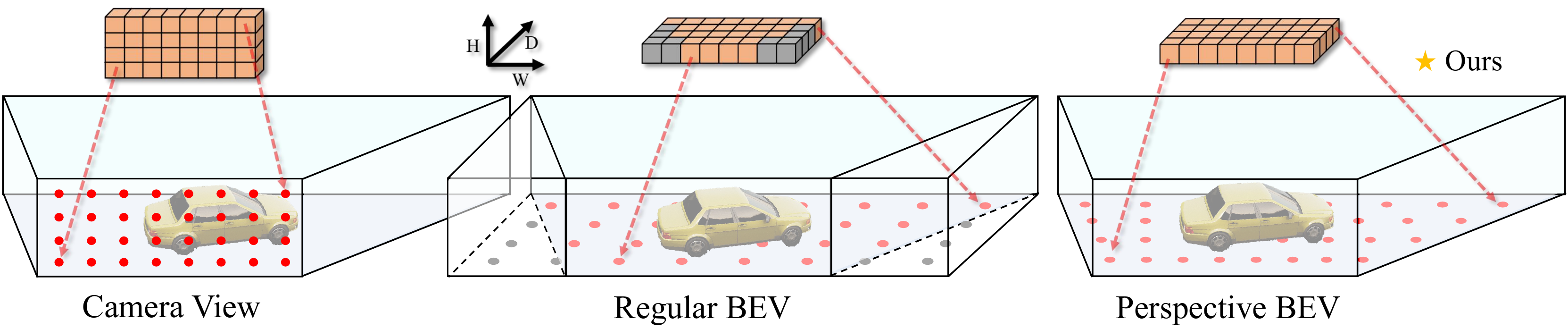}
    \caption{Different features and corresponding anchor distributions for CV detectors, regular BEV detectors and PersDet. The red dots denote anchor points inside FOV, gray dots denote invalid anchors outside FOV.}
    \label{fig:anchors}
\end{figure*}

\begin{abstract}
Currently, detecting 3D objects in Bird's-Eye-View (BEV) is superior to other 3D detectors for autonomous driving and robotics. However, transforming image features into BEV necessitates special operators to conduct feature sampling. These operators are not supported on many edge devices, bringing extra obstacles when deploying detectors.
To address this problem, we revisit the generation of BEV representation and propose detecting objects in perspective BEV --- a new BEV representation that does not require feature sampling. We demonstrate that perspective BEV features can likewise enjoy the benefits of the BEV paradigm. Moreover, the perspective BEV improves detection performance by addressing issues caused by feature sampling. We propose PersDet for high-performance object detection in perspective BEV space based on this discovery. While implementing a simple and memory-efficient structure, PersDet outperforms existing state-of-the-art monocular methods on the nuScenes benchmark, reaching 34.6\% mAP and 40.8\% NDS when using ResNet-50 as the backbone.
\end{abstract}

\section{Introduction}
3D object detection is pivotal in helping intelligent agents perceive the environment in autonomous driving systems and robotics. Sensors like cameras~\cite{monodet,3dbbox,monodet2}, LiDARs~\cite{ku2018joint,pvrcnn,pointpillars}, and RaDARs~\cite{voxelnet,pointrcnn} are broadly used for this task, among which camera-based methods have drawn growing attention due to their lower cost.

According to the view that detection is conducted, camera-based 3D object detectors can be categorized into two major classes: Camera-View (CV) detectors and Bird's-Eye-View (BEV) detectors. 
CV detectors, such as FCOS3D~\cite{fcos3d} and PGD~\cite{pgd}, follow the design of 2D detectors, extend the detection head for 3D tasks. These methods often show inferior performance to BEV methods, according to the nuScenes leaderboard. However, since CV detectors directly detect objects on images, they can retain a fully convolutional structure just like 2D detectors. This character makes them deploy-friendly and thus popular in the industry. 

In contrast, detecting 3D objects in BEV~\cite{caddn,bevdet,bevdet4d} is a new paradigm designed for 3D scenarios, showing excellent performance in 3D tasks. To conduct detection in BEV space, they use a view transforming stage that transforms features from Camera-View into Bird's-Eye-View. An extended 2D detecting head~\cite{centerpoint} is then applied to the transformed BEV features.
Among existing transforming methods, Lift-Splat~\cite{lss} based methods have drawn growing attention due to their high efficiency and efficacy. 
The Lift-Splat-based~\cite{lss} methods transform views in three steps: 1) Projecting 2D image features to 3D space using estimated depth. This step produces a 3D representation that allows one to view features from BEV. 2) Aligning 3D image features to pre-defined BEV grids/anchors. These anchors are often distributed regularly (\emph{i.e.}, uniformly) in space. Since the 3D representation carries perspective distortion derived from perspective images, this step adopts \emph{feature sampling} operation to align features with the BEV anchors and remove the distortion. 3) Collapse the 3D feature along height dimension to obtain the 2D BEV feature.
While other steps in the Lift-Splat transformation show a simple pipeline, the second step adopts the feature sampling operator that can not be deployed or accelerated on many devices, being the major impediment to making BEV detectors feasible in industrial scenarios. This situation makes us wonder: 
\emph{Can we enjoy the benefits of the BEV paradigm while retaining a simple pipeline?}

To answer this question, we delve into the details of the alignment between features and anchors. As demonstrated in Fig.~\ref{fig:anchors}, CV detectors lay anchors in the image plain along the height and width axes. These anchors are naturally aligned to the down-sampled feature map extracted from the image. In contrast, the anchors of BEV detectors distribute among the width and depth axes since the semantics of the BEV feature correspond to objects on the horizontal reference plane. Since the Lift-Splat method is initially designed for segmentation tasks, these anchors conventionally distribute uniformly on the reference plane~\cite{lss}.
Such distribution introduces the requirement of feature sampling --- the perspective effect is carried in the projected feature but not in the anchors. In this paper, we propose introducing the perspective effect to the anchors rather than removing it from the features. In that case, the semantic information can be aligned with real-world anchors without feature sampling. Furthermore, since the feature sampling operations in existing methods often accompany by information loss and structural distortion (see Fig.~\ref{fig:sampling}) caused by over-sampling and under-sampling phenomena, the removal of feature sampling results in extra performance gains.

With this improvement, we build a high-performance detector that directly conducts detection on perspective BEV features --- PersDet. 
Concretely, we first project image features into the perspective BEV space following a Lift-Splat~\cite{lss} like pattern. Then, instead of sampling the features, we directly collapse this 3D representation and put our PersHead onto the BEV features with perspective distortion. The PersHead deploys anchors for regression in the perspective space, where anchors are sparse in the remote area and dense on the contrary. In that case, the feature produced by outer production is naturally aligned to anchors.

We conclude our contributions as follows:
\begin{enumerate}
    \item We show the drawbacks of feature sampling and propose perspective BEV to avoid this process.
    \item We propose PersDet, which performs object detection on perspective BEV features, validating the effectiveness of perspective BEV features.
    \item The proposed PersDet achieved state-of-the-art performance on the nuScenes benchmark while showing the advantage of being deploy-friendly.
\end{enumerate}

\section{Related Works}
\subsection{Camera-View 3D Object Detection}
Image data is the most convenient and economical to obtain among data forms in autonomous driving and robotics. Image-based 3D object detectors use image data and associated camera parameters as input to detect objects within the image range and predict their 3D bounding boxes. 
Camera-View detectors like FCOS3D~\cite{fcos3d} and PGD~\cite{pgd}, inspired by 2D object detector FCOS~\cite{fcos}, extend the FCOS detection head for extra bounding box attributes and depth prediction. These detectors retain a fully convolutional structure and show fair performance on several benchmarks~\cite{kitti,nuscenes} but have the problem of perceiving target translation. 
SS3D~\cite{ss3d} proposes a single-stage monocular 3D object detector that produces a redundant representation of each relevant object in the image together with uncertainty estimations. MonoDIS~\cite{monoDis} presents decoupled regression losses to better multi-task training. Our work, using a single image as input, differs from previous work that extracts features on the image view. We project the features onto the BEV space and perform 3D object detection on the BEV space. 
PGD propose using geometric constraints and depth probabilities to improve depth estimation accuracy base on FCOS3D. It considerably reduces the depth estimation problem while increasing the compute budget and inference delay. 

\subsection{BEV Representation}
The Bird's Eye View (BEV) representation was firstly proposed for segmentation task~\cite{vpn,pon,lss} in 3D scenarios, in which models aim to predict surrounding maps from images. These models use an image encoder to obtain the image feature and then apply view-transformation to get the BEV feature. The transformation can be implemented by data-driven methods or depth-based methods. 

Among data-driven methods, VPN~\cite{vpn} uses MLP to transform flatten CV features into BEV features. PON~\cite{pon} uses column-wise (rather than image-wise) MLP to perform transformation on each direction. These methods efficiently implement transformation but perform poorly in detection and segmentation tasks. 

For depth-based methods, they often outer product the image feature and categorical depth distribution to obtain BEV features. OFT~\cite{oft} uses a uniformly distributed depth to perform projection.
LSS~\cite{lss} predicts categorical depth distribution in an unsupervised way, using 3D bounding boxes for implicit depth supervision. CaDDN~\cite{caddn} introduces depth supervision from LiDAR to obtain high-quality depth prediction.
These methods efficiently generate high-quality BEV features, but they all apply feature sampling methods like grid sampling~\cite{caddn} or voxel pooling~\cite{bevdepth} on the projected BEV feature to obtain a non-perspective BEV feature. That is because previous works generate BEV features for segmentation tasks, where a canonical feature is required. This operation becomes the major impediment to deploying and accelerating these models.

\subsection{BEV 3D Object Detection}
Based on LSS, BEVDet~\cite{bevdet} introduces the detection head to the generated BEV feature and implements object detection in BEV space. Unlike BEVDet, CaDDN~\cite{caddn} generates a BEV feature map for each camera rather than a single BEV feature for each scene. BEVDepth~\cite{bevdepth} introduces depth supervision based on BEVDet. This work also proposes an individual DepthNet for better depth prediction. BEVDet4D~\cite{bevdet4d} propose concatenating multi-frame BEV features to predict object velocity better.

This work explores the detection capability of the ``uncorrected'' BEV feature. By implementing object detection on the perspective BEV feature, we eliminate any complex operation and propose a fully convolutional BEV detector.

\begin{figure}[]
	\centering
    \includegraphics[width=\columnwidth]{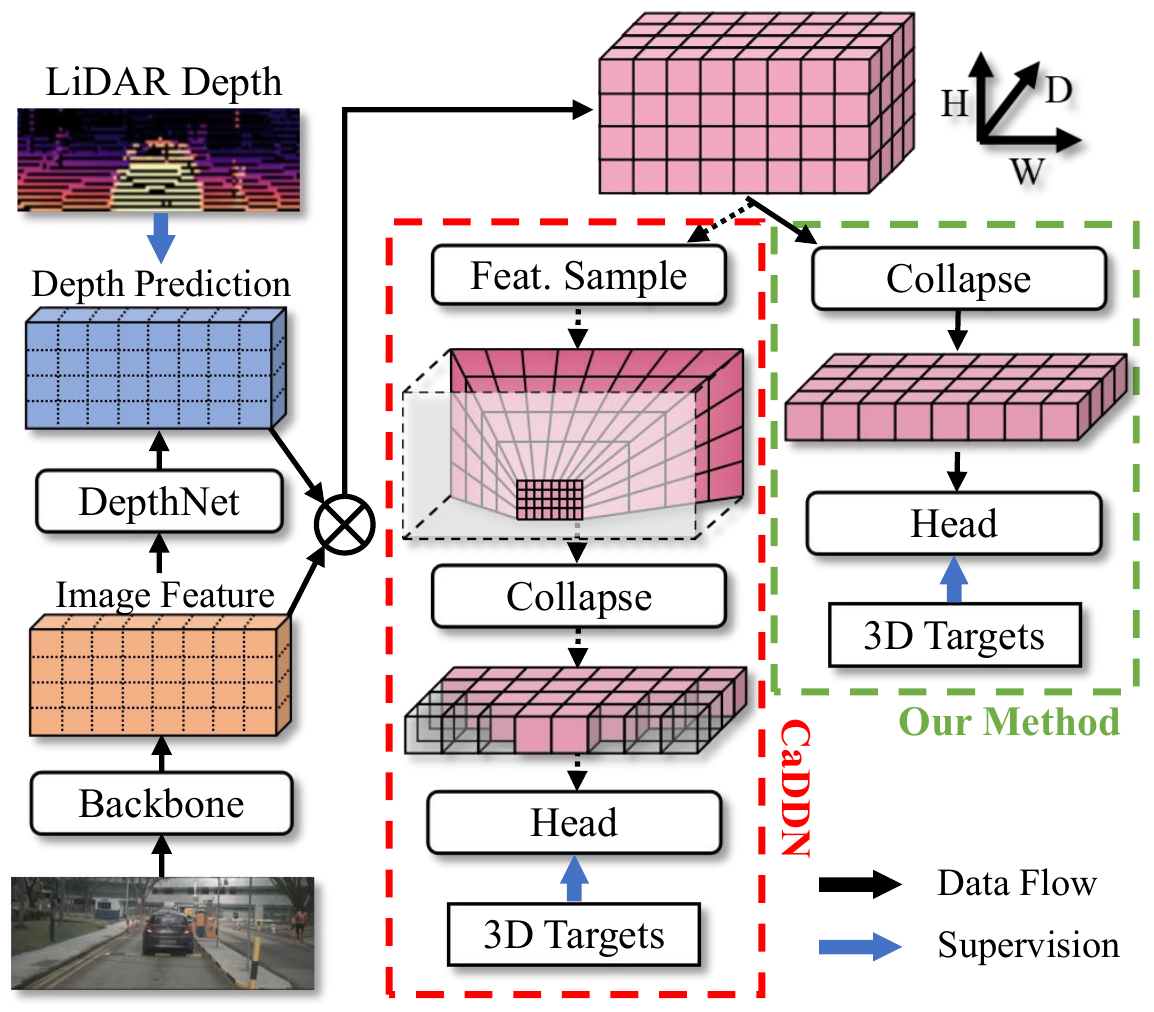}
    \caption{Illustration of existing BEV framework (CaDDN) and our improvement.  Lift-Splat-based methods adopt feature sampling before collapsing the 3D feature (red), and we propose to remove this procedure to produce a better BEV representation (green).}
    \label{fig:dloss}
\end{figure}

\section{PersDet}
In this section, we first introduce the pipeline of existing monocular BEV detectors in Sec~\ref{sec:overall}. Then, in Sec.~\ref{sec:oversample} we analyze the disadvantages of feature sampling adopted in existing methods. To overcome these problems, we propose PersDet as our solution in the remaining part, detailed information, including structural design and learning target design, is introduced.
\subsection{BEV Detection Framework}
\label{sec:overall}

Existing BEV detectors represented by CaDDN~\cite{caddn} follow a Lift-splat pattern, in which a feature extractor and a DepthNet are adopted for feature extraction and depth prediction. The depth prediction is individually supervised by the ground truth depth obtained by LiDAR. Note that the LiDAR depth is only utilized during training; the detector takes only images for testing and validation. 
As illustrated in Fig.~\ref{fig:dloss}, the BEV detection framework performs uniformed training using both 3D bounding boxes and depth supervision. The overall loss $\mathcal{L}$ of PersDet is defined as:
\begin{equation}
    \mathcal{L} = \mathcal{L}_{det}+w_d \mathcal{L}_{d}
\end{equation}
where $\mathcal{L}_{det}$ is the detection loss, $\mathcal{L}_{obj}$ and $w_d$ is the depth loss and its weight.

With the extracted image features $F\in \mathbb{R}^{C\times H\times W}$ and depths $D\in \mathbb{R}^{D\times H\times W}$, the projection is performed by doing the outer product of these tensors. 
Traditionally, the generated 3D features $F_{3D}\in \mathbb{R}^{C\times D\times H\times W}$ is then sampled or pooled to align with regular anchors and remove the perspective effect. A CollapseConv is used after feature sampling to obtain the 2D BEV feature representation $F_{BEV}\in \mathbb{R}^{C\times D\times W}$:
\begin{equation}
    F_{3D} = F \otimes D
\end{equation}
\begin{equation}
    F_{3D}^* = GridSample(F_{3D})
\end{equation}
\begin{equation}
    F_{BEV} = CollapseConv(F_{3D}^*)
\end{equation}

However, in the below section, we show that the feature sampling operation can bring several disadvantages and thus be sub-optimal.

\subsection{Disadvantages of Feature Sampling}
In this section, we analyze the behavior of feature sampling. We take CaDDN as a representative monocular BEV detector, which adopt Grid Sampling to align BEV feature.
\subsubsection{Over-Sampling and Under-Sampling}
\label{sec:oversample}
\begin{figure}[]
	\centering
    \includegraphics[width=0.95\columnwidth]{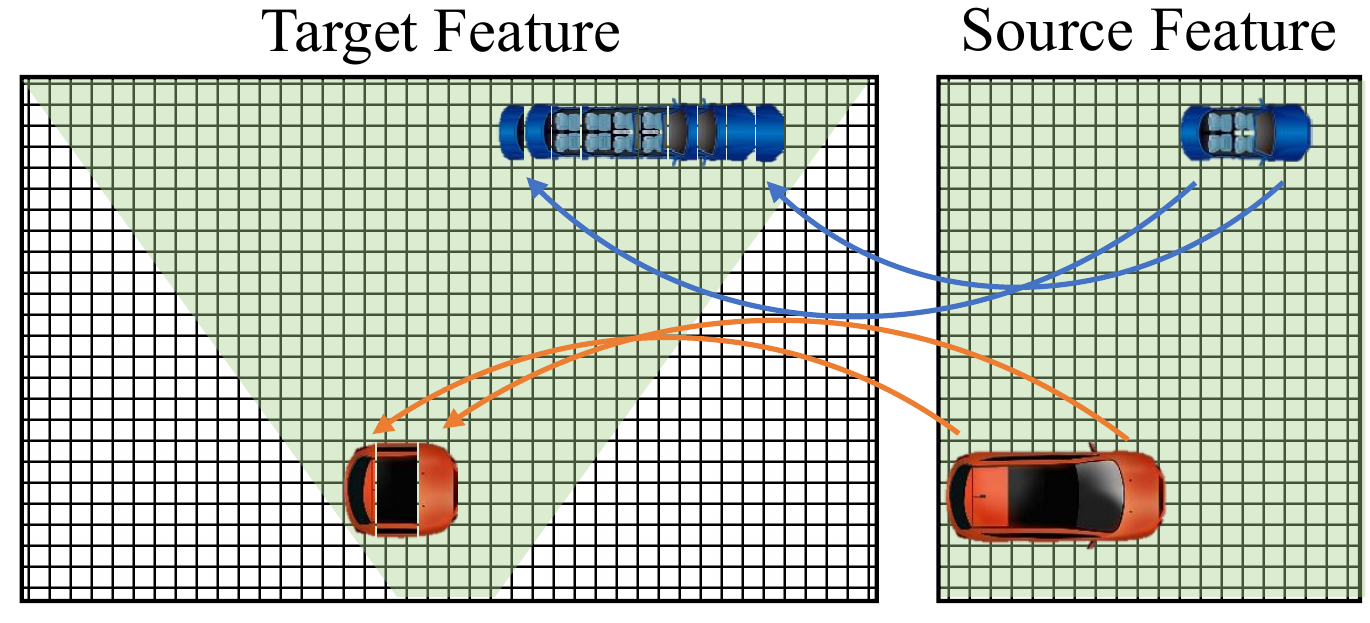}
    \caption{Illustration of over-sampling (blue car) and under-sampling (orange car) effect. Inner semantic information of object is destroyed due to feature sampling.}
    \label{fig:sampling}
\end{figure}
The Grid Sampling~\cite{caddn} operation samples features based on spatial coordinate information and pre-defined grids (anchors). The anchors' setting has a decisive impact on the performance and efficiency of this process. As illustrated in Fig.~\ref{fig:sampling}, when giving a dense anchor distribution, the problem of over-sampling could result in duplicated feature representation. The over-sampling phenomenon will cause a waste of memory and deterioration of structural integrity. In contrast, a sparse distribution could cause an under-sampling problem where some source features are not sampled, resulting in information loss. 
Worse still, the over-sampling and the under-sampling problem can not be solved entirely since they always occur simultaneously. Because the perspective camera FOV has different widths at different depths (wide in the distance and narrow on the contrary), there is always an under-sampling problem in the near field and an over-sampling problem at a distance. The best performance obtained by adjusting the hyper-parameter is only a trade-off.
Therefore, the deterioration of features is inevitable once feature sampling is adopted.
\begin{figure}[]
	\centering
    \includegraphics[width=0.85\columnwidth]{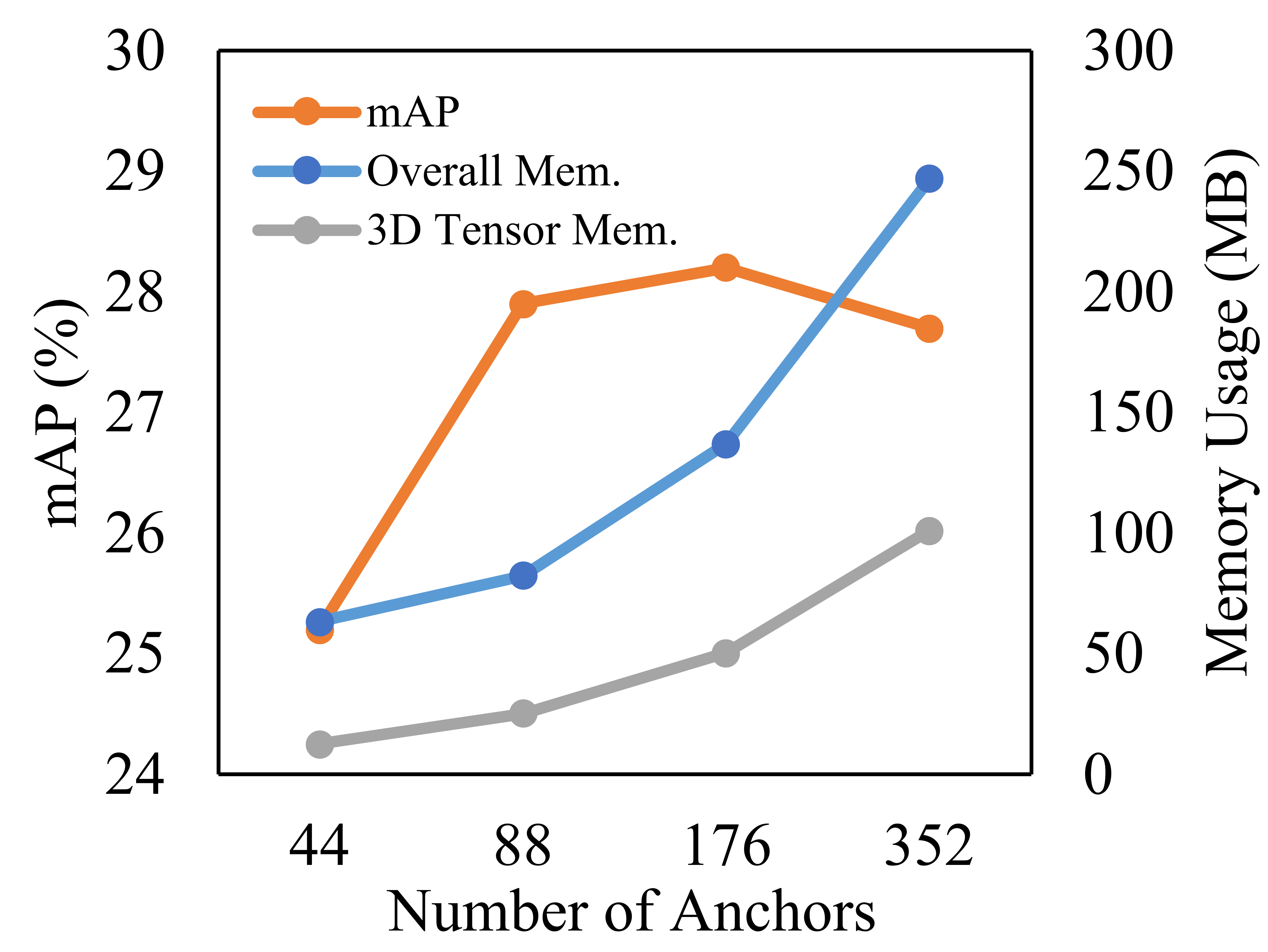}
    \caption{The impact of increasing (horizontal) anchor density on performance and memory consumption. }
    \label{fig:mem}
\end{figure}

\subsubsection{Memory Inefficiency}
The irregular camera Field Of View (FOV, the green region in Fig.~\ref{fig:sampling}) causes the memory inefficiency of the regular BEV feature. Since computing devices store and calculate data in the form of matrices, and the FOV of the camera is a frustum, there is a large portion of the memory area that does not contain valid information. Such inefficiency leads to a dramatic increase in the size of the 3D tensor, accompanied by significant memory waste. 
In Fig.~\ref{fig:mem} we present the performance and memory consumption of the CaDDN detector under different anchor densities on the X-axis (horizontal axis of images). It shows that the memory consumption of the 3D tensor grows quadratically with increasing anchor density and leads to increasing overall memory consumption. When the number of anchors increases to 176, the under-sampling problem is alleviated, allowing an adequate performance. However, the expansion of the 3D tensor has doubled the overall memory consumption of the detector.
In other words, to get a suitable performance, the feature sampling operation could significantly increase the memory consumption of detectors, leading to higher model application costs.

\subsubsection{Troublesome Operator}
As mentioned in the above sections, the feature sampling uses custom operations like Grid Sampling or Voxel Pooling~\cite{bevdepth}. These operations can be deployed and accelerated by employing custom operators on flexible platforms represented by CUDA. However, there are also many other platforms on edge devices where a such complex operator can not be performed or accelerated. Therefore, the advantages of the BEV detector cannot be enjoyed by many low-end scenes due to the feature sampling.

Considering the above shortcomes of feature sampling, we explore performing detection directly on projected BEV features where the perspective effect has not been removed. The PersHead, as the solution to the above problems, is proposed for simple and high-performance object detection.

\subsection{Object Detection in Pespective BEV Space}
\begin{figure}[]
	\centering
    \includegraphics[width=\columnwidth]{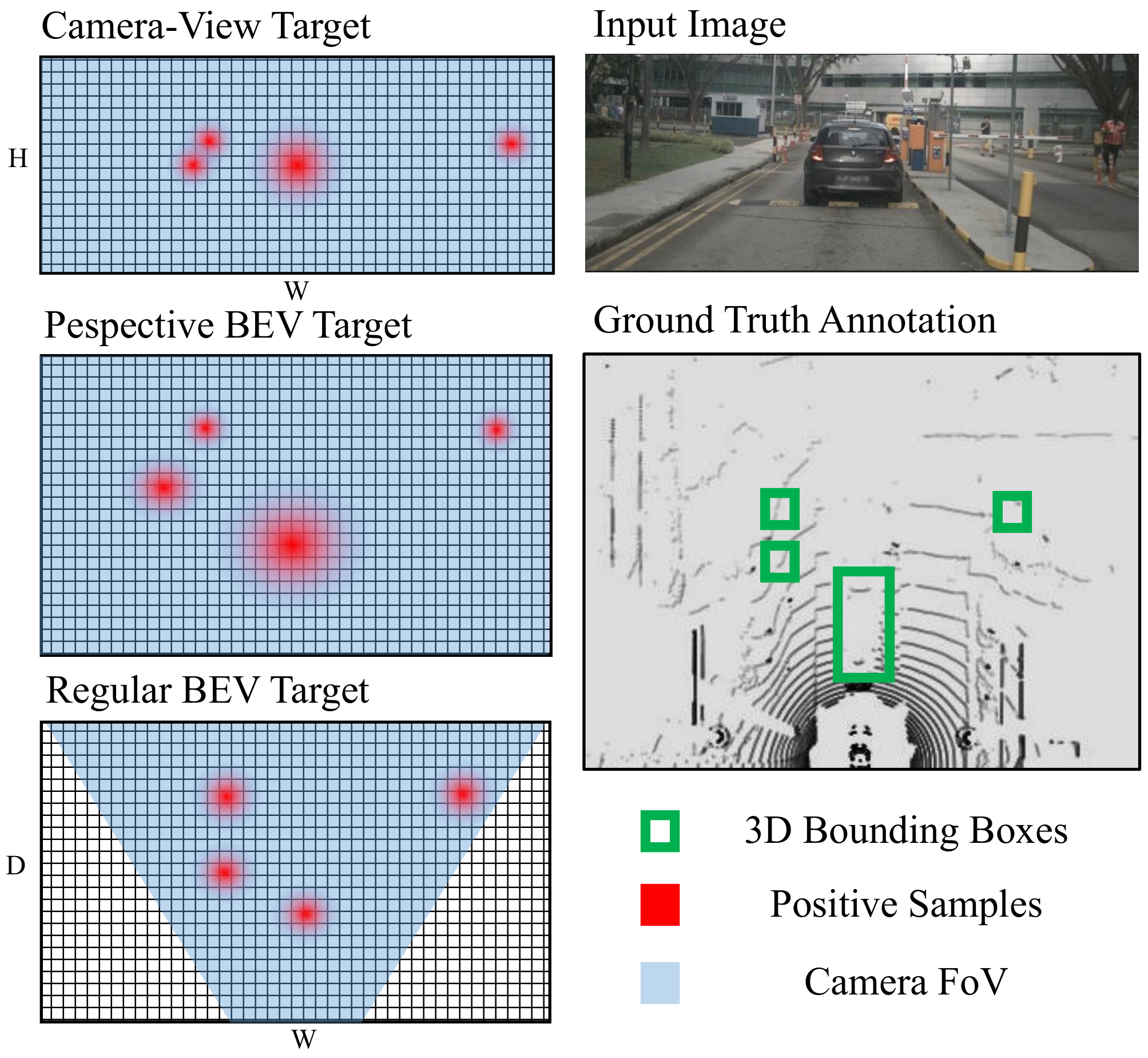}
    \caption{Heatmap target of regular BEV detectors, CV detectors and the PersDet.}
    \label{fig:head}
\end{figure}
\subsubsection{Anchors and Targets}
We design the perspective anchors to align with the perspective BEV features. Specifically, we apply inverse projection to the uniformly distributed \emph{frustum} anchors to obtain anchors in real-world coordinates. Giving frustum anchors $A^{frustum}$, the inverse projection is applied as:
\begin{equation}
    A^{world*} = I^{-1} \times A^{frustum}
\end{equation}
where $I$ is the camera intrinsic matrix. After this inverse projection, the distribution of perspective anchors is sparse in the distance and dense on the contrary.
Since these anchor points contain a perspective effect, corresponding changes are required for the heatmap target (\emph{i.e.}, objectness score targets). As shown in Fig.~\ref{fig:head}, the perspective effect is introduced to the target to align with perspective features. 

\subsubsection{PersHead}
We design the PersHead following the CenterPoint head~\cite{centerpoint}, using task-specific branches for classification and regression. Giving BEV feature $F^{bev} \in \mathbb{R}^{C\times W\times D}$ produced by the Lift-Splat module, the PersHead performs shared convolution on the BEV feature and makes predictions using task-specific sub-nets in a dense form. For each grid of the BEV feature $F^{bev}_{w,d}$, the PersHead predicts its objectness score $P^{obj}_{w,d}$ and bounding box attributes $P^{attr}_{w,d}$ including center offsets, center height, box size, box local yaw~\cite{fcos3d}, direction category, and velocity. The overall detection loss $\mathcal{L}_{det}$ is calculated as:
\begin{equation}
    \mathcal{L}_{det} = CELoss(P^{obj},T^{obj}) + L1Loss(P^{attr},T^{attr})
\end{equation}
where $T^{obj}_{w,d}$ is the objectness score target and $T^{attr}$ is the attributes of boundingbox on each grid.

Finally, with the predicted box attributes, the real coordinates of the box center can be obtained by adding the center offsets to the coordinates of corresponding anchors.

\begin{table*}[!h]
	\begin{center}
		\begin{tabular}{c|cc|ccccccc}
  \toprule
	Methods  & Image Size & mAP & mATE & mASE & mAOE & mAVE & mAAE & NDS\\
\midrule
CaDDN~\cite{caddn}
    & $704\times256$ &0.294 &0.702 &0.283 &0.579 &0.988 &0.222 &0.370\\
BEVDet§~\cite{bevdet}
    & $704\times256$ &0.298 &0.725 &0.279 &0.589 &0.860 &0.245 &0.379\\
PersDet
    & $704\times256$ &\textbf{0.319} &0.676 &0.284 &0.589 &0.924 &0.229 &\textbf{0.389}\\

\midrule
FCOS3D*~\cite{fcos3d}
    & $1600\times900$&0.288 &0.777 &0.266 &0.544 &1.228 &0.170 &0.368\\
PGD*~\cite{pgd}
    & $1600\times900$&0.320 &0.735 &0.266 &0.492 &1.114 &0.170 &0.394\\
PersDet
    & $1408\times512$&\textbf{0.346} &0.660 &0.279 &0.540 &0.964 &0.207 &\textbf{0.408}\\
\bottomrule
  \end{tabular}
\caption{Experimental results on nuScenes. All experiments use \emph{ResNet-50} as backbone. Detectors with * are trained under settings from original implementation to get best performance. The method with § uses multi-view fusion.}
\label{table:sota}
	\end{center}
\end{table*}

\section{Experiments}

\subsection{Implementation Details}
\subsubsection{Dataset}
We conduct experiments on the nuScenes~\cite{nuscenes} dataset, which is a challenging autonomous driving benchmark.
For the detection task, 1.4M annotated bounding boxes from nuScenes are categorized into six meta-classes (tasks) and ten classes. Following previous works~\cite{bevdet,bevdepth}, we only detect and evaluate objects within 58 meters. We report the evaluation metrics defined in the official nuScenes API, including mAP, ATE, ASE, AOE, AVE, AAE, and the NuScenes Detection Score (NDS). We mainly focus on the mAP and the NDS metrics for monocular detection. The mAP is evaluated as the 2D mAP on the ground plane. The NDS is a reweighted combination of other metrics, indicating an overall detecting capability.

\subsubsection{Training Settings}
Our experiments are implemented on PyTorch~\cite{pytorch} with CUDA acceleration. If not otherwise specified, we use ResNet-50~\cite{resnet} as our backbone, and a SECOND FPN~\cite{second} is adopted to merge multi-level image features. Following BEV detectors that report performances on nuScenes~\cite{bevdet,bevdepth}, most of our experiments were performed under $256\times 704$ resolution with batchsize 8 ($8\times 6$ images for six images per sample). Data augmentation, including RandomFlip, Random Rotate, Random Scaling, and Random Cropping as in BEVDepth~\cite{bevdepth} are used. We train our models on the nuScenes dataset for 24 epochs, using learning rate 2e-4, depth weight 3, and EMA strategy. When comparing with other methods on nuScenes, CBGS~\cite{cbgs} is adopted.

\subsubsection{Reproduction of CaDDN}
In order to conduct a fair comparison, we reproduce CaDDN~\cite{caddn} on our code base. We align the backbone, FPN, task-specific head, image feature size, and the training setting of the reproduced CaDDN and our model. 
For the new nuScenes dataset, we use the voxel grid range $[2,58]\times [-40,40] \times [-5,3]$ (m) and voxel grid size $[0.64, 0.64, 0.64]$ (m) since our feature down-sample rate is 16.

\subsection{Comparison with State-of-the-Arts}
\label{sota}
We compare our method with both monocular and multi-ocular detectors since they adopt a similar pipeline. For multi-ocular detectors, BEVDet~\cite{bevdet} is taken as a representative method.
As demonstrated in Tab.~\ref{table:sota}, PersDet outperforms existing monocular detectors with significant advantages. 
Besides, thanks to the explicit modeling of depth and depth supervision, PersDet shows a greater performance over the multi-ocular detector --- BEVDet by a margin of 2.2\% mAP. 
When giving high-resolution images as input to compare with FCOS3D~\cite{fcos3d} and PGD~\cite{pgd}, PersDet also shows better performance even when taking images of lower resolution (for simplicity, we scale the inputs by an integer multiple).

\begin{table*}[!h]
	\begin{center}
		\scalebox{0.95}{
		    \begin{tabular}{c|c|c|c|ccccccc}
\toprule
View & Depth / Depth Sup. & Feat. Sample & Feat. Size & mAP & mATE & mASE & mAOE & mAVE & mAAE & NDS  \\
\midrule
CV & bboxes & - & $44\times 16$
& 0.189 & 0.925 & 0.293 & 0.636 & 1.056 & 0.269 & 0.282 \\
\midrule
BEV & bboxes \& LiDAR & GS  & $44\times56$ 
& 0.252 & 0.838 & 0.295 & 0.708 & 1.087 & 0.269 & 0.315 \\
BEV & bboxes \& LiDAR & GS  & $125\times87$ 
& 0.276 & 0.758 & 0.290 & 0.658 & 1.042 & 0.260 & 0.341 \\
BEV & bboxes \& LiDAR & VP  & $44\times56$ 
& 0.245 & 0.914 & 0.302 & 0.695 & 0.992 & 0.268 & 0.306 \\
BEV & bboxes \& LiDAR & VP  & $125\times87$ 
& 0.275 & 0.766 & 0.293 & 0.657 & 1.075 & 0.257 & 0.340 \\
\midrule
BEV & bboxes \& LiDAR & $\times $ & $44\times56$ 
& \textbf{0.300} & 0.735 & 0.296 & 0.678 & 1.104 & 0.312 & \textbf{0.348} \\
BEV & bboxes & $\times $ & $44\times56$ 
& 0.290 & 0.795 & 0.289 & 0.685 & 1.068 & 0.293 & 0.339 \\
BEV & static random & $\times $ & $44\times56$ 
& 0.249 & 0.849 & 0.293 & 0.722 & 1.100 & 0.325 & 0.306 \\
BEV & uniform & $\times $ & $44\times56$ 
& 0.052 & 0.908 & 0.306 & 0.739 & 1.099 & 0.323 & 0.200 \\

\bottomrule
            \end{tabular}
}
    \caption{Performances of detectors in different views and sample densities. GS stands for GridSampling, and VP stands for VoxelPooling. Models were trained without CBGS.}
    \label{table:fa}
    \end{center}
\end{table*}

\subsection{Boosts of BEV Feature}
\label{analy:anchors}
In order to sort out the boost of PersDet from the BEV paradigm, we conduct a series of experiments to compare detectors with different feature distributions. We use detectors with the same backbone, same FPN, and same head structure to fairly compare the impact of feature distribution. 
For the Camera-View detector, we remove the view transformer and let the detector predict object depth using an individual branch. 

\subsubsection{Features on Depth Dimension}

\begin{figure}[]
	\centering
    \includegraphics[width=0.95\columnwidth]{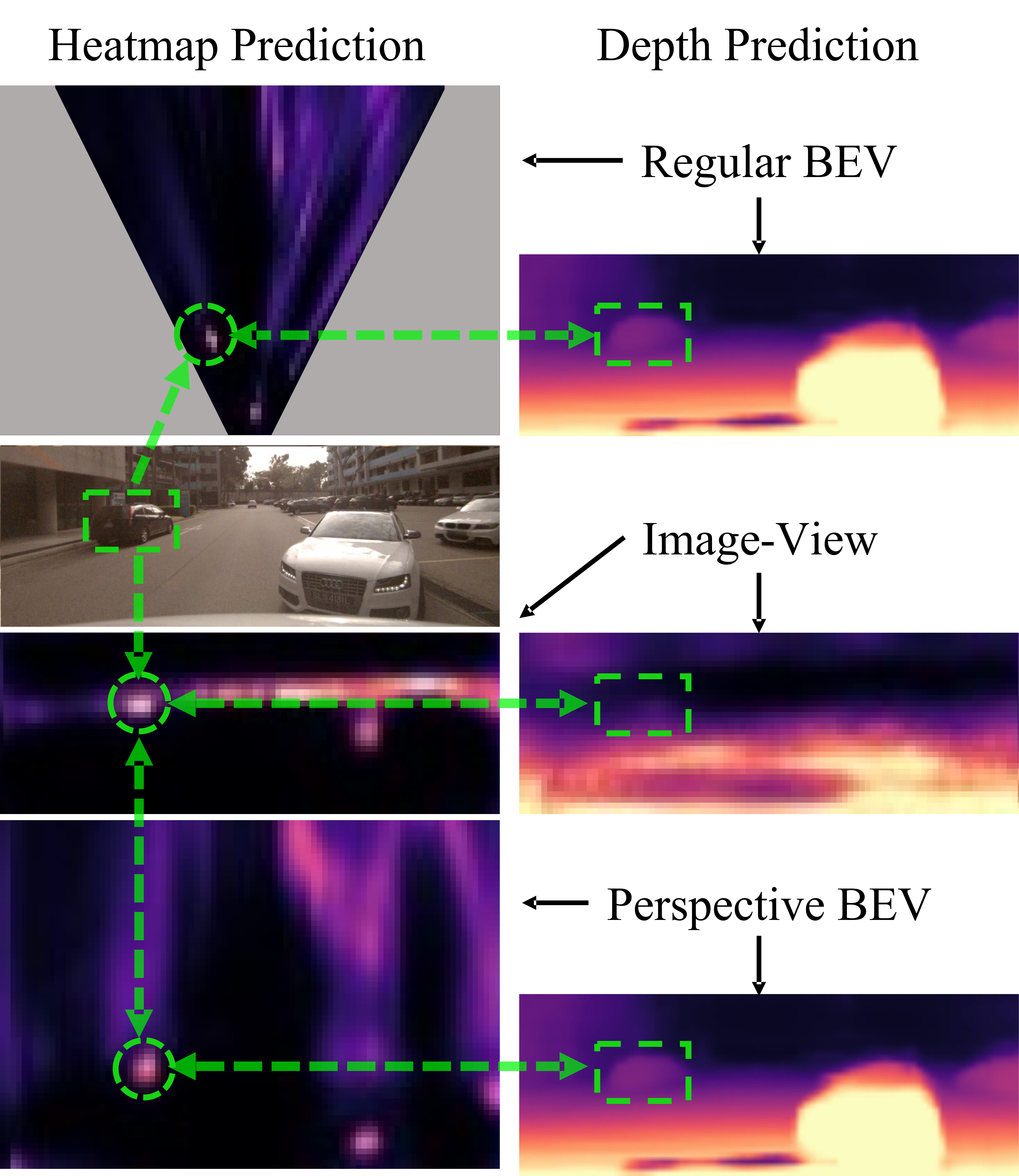}
    \caption{Visualization of heatmap and depth predictions of different detctors. The heatmap of regular BEV detector contains a large portion of invalid area (gray area).}
    \label{fig:depth}
\end{figure}

To demonstrate the effect of feature distribution, we first replace the categorical depth prediction of PersDet with a random tensor (static during training and evaluation) to remove the boost of any depth information. 
Image features are projected into BEV space with random intensity with the static random depth distribution. 
In this case, the PersDet decreased 4.1\% mAP, still showing a gap of 5\% mAP with its CV counterpart. 
We then set the categorical depth of PersDet to uniformly 1. Since the BEV feature is the very same at any depth, the feature distribution along the depth dimension is counterbalanced. Under this setting, the PersDet shows really poor performance and gets to 5.2\% mAP, even lower than the CV detector. These results show that the BEV detector obtains most of the performance boost by distributing features along the depth dimension. Since the mAP metric in autonomous driving does not require precise height, the locating ability of detectors is dependent on horizontal and depth-wise perception. By modeling depth positions and horizontal positions into a unified heatmap, BEV detectors learn to locate objects in an end-to-end pattern, thus leading to better performance.

\subsubsection{Precise Depth Prediction}
We also note that the differences between the CV and BEV detectors are mainly represented in the ATE (Average Transition Error) and mAP metrics. 
Considering the fancy performance of CV detectors in 2D tasks, it is fair to attribute such low performance to imprecise depth.
While the CV detector predicts object depth using an individual branch, the BEV representation transforms the depth prediction from a regression task into a classification task, learning depth in a joint pattern. Such a joint training scheme enables the detector to produce depth fit for the detection task. Furthermore, the experimental results show that the extra depth supervision during training could boost 1\% mAP and 1\% NDS.
The improvement of depth prediction can also be seen in the depth prediction visualization (see right of Fig.~\ref{fig:depth}), in which BEV detectors clearly address the target car in depth prediction, while the CV detector fails to sort out the car from the background.

\subsection{Information Loss of Feature Sampling}
\label{analy:scale}
As stated in Sec.~\ref{sec:oversample}, the performance of CaDDN is affected by voxel-grid resolution due to under-sample and over-sample problems. 
In Sec.~\ref{sota}, we follow the original implementation of CaDDN, use fixed grid size and obtains a feature size (width$\times$depth) of $125\times87$. In this section, we scale this feature to a different shape to demonstrate its effect on performance.
It can be seen from Tab.~\ref{table:fa} and Tab.~\ref{table:scale} that as the density of the sample grid increases, the model performance firstly shows an observable boost. However, when the sampling grid is too dense, the detector performance shows a significant drop. This phenomenon validates our under-sampling and over-sampling analysis, indicating that feature sampling methods are sensitive to this hyper-parameter, bringing extra obstacles when transplanting the model to a new area.

Furthermore, even when CaDDN obtains the best performance, there is still a significant performance drop caused by feature sampling.
As seen in Tab.~\ref{table:fa} and Tab.~\ref{table:scale}, the PersDet shows a boost of 1.4\% mAP compared with the best performance of CaDDN. We attribute this gain to the efficient and high-quality perspective BEV features.

\subsection{Latency Improvements}

\begin{figure}[h]
	\centering
	\begin{tabular}{@{}c@{}c@{}}
	\includegraphics[width=0.5\columnwidth]{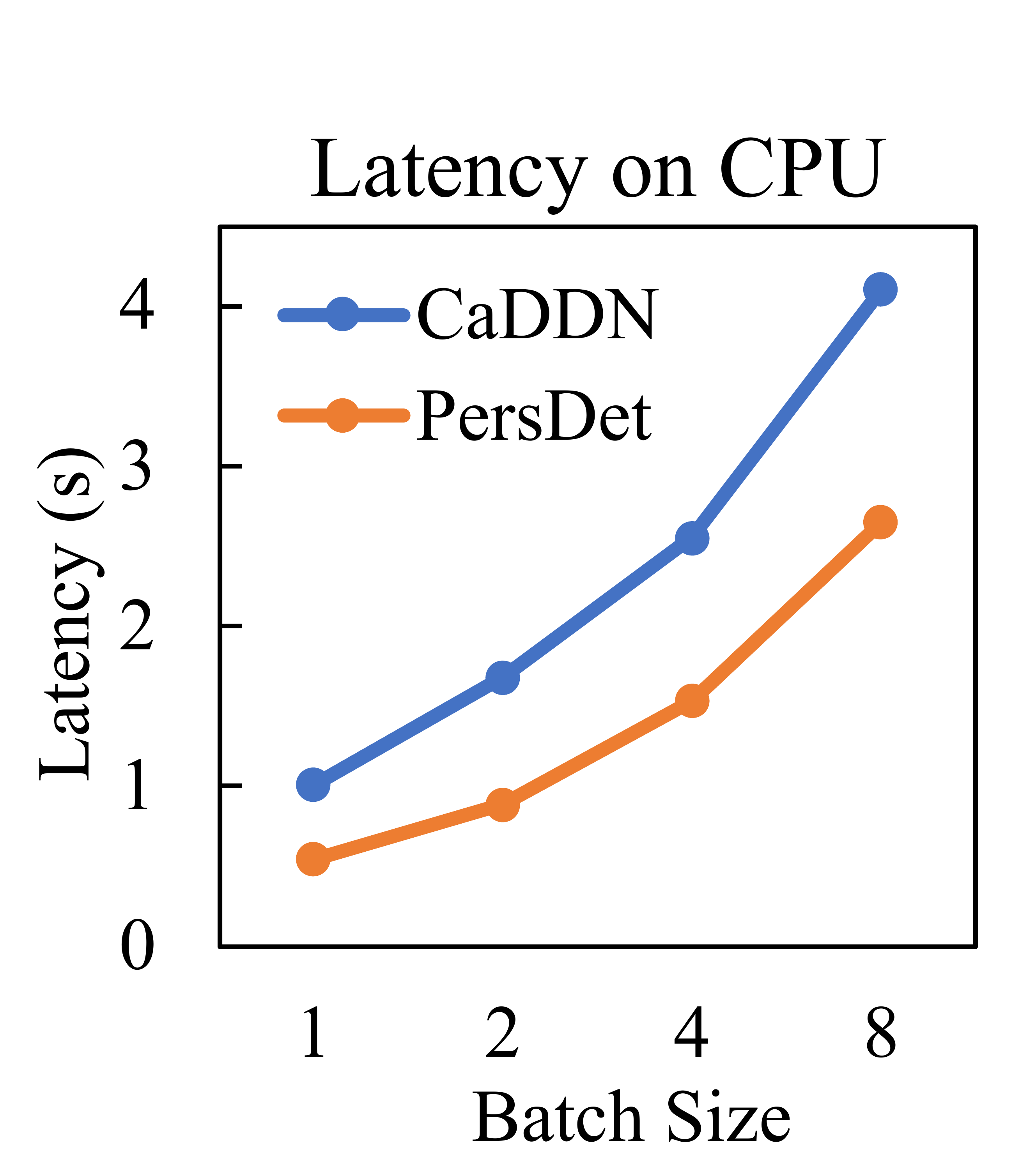} &  \includegraphics[width=0.5\columnwidth]{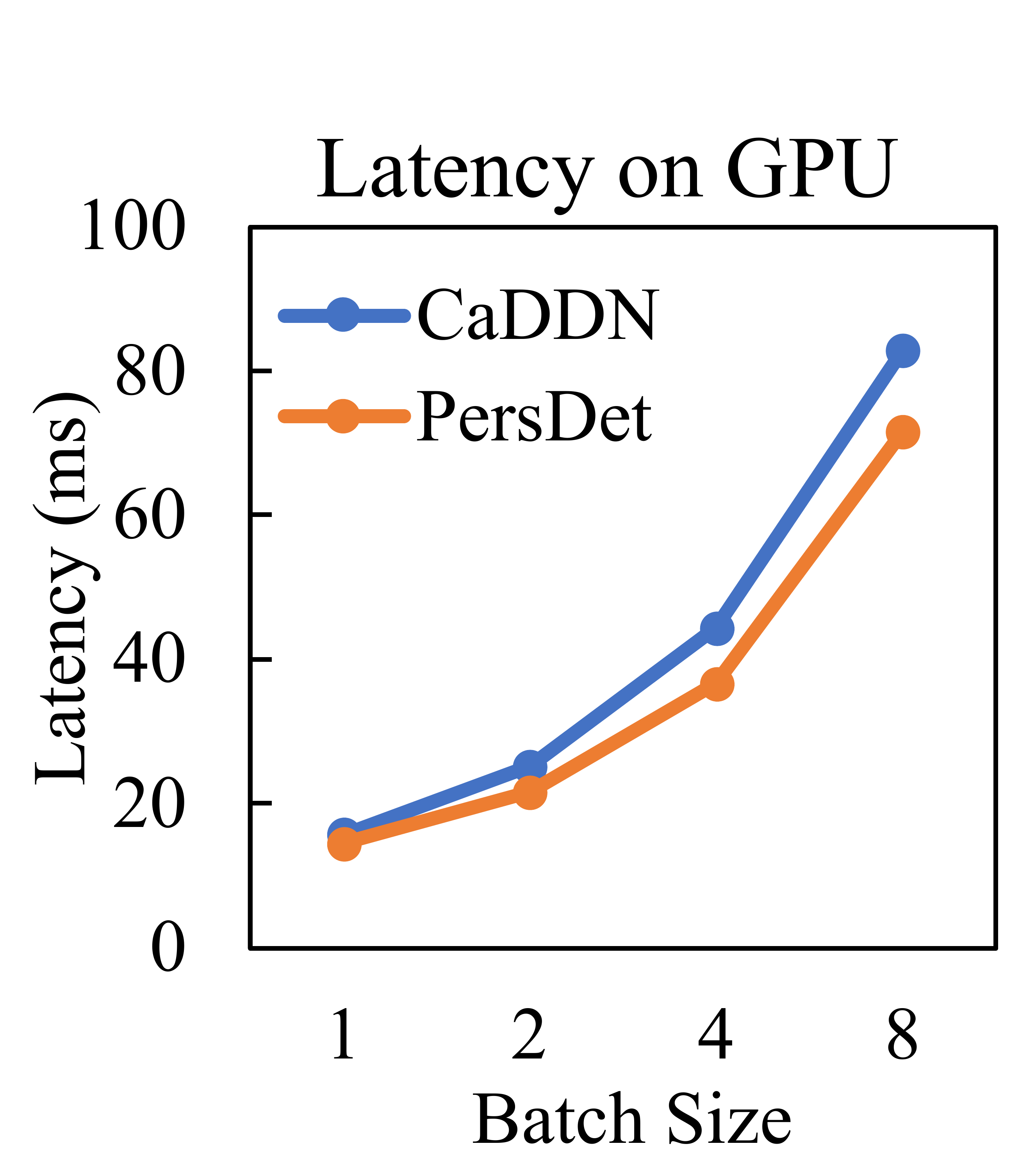}
	\end{tabular}
	\caption{Latency of CaDDN and PersDet on different devices.}
	\label{fig:speed}
\end{figure}

By removing the feature sampling, PersDet improves its inference speed by a large margin. We measure the inference speed on GPU with CUDA acceleration and CPU without any acceleration in Fig.~\ref{fig:speed}. The CaDDN takes a sampling density of $44\times 56\times 16$, the same as the source feature. The latency data indicate that PersDet has a significant advantage in speed over methods using Grid Sampling~\cite{caddn}. On CPU platforms without custom operators (like the case of edge devices), the latency of PersDet is reduced by 46\% for a single image (batch 1). Although CUDA accelerates the feature sampling on the GPU platform, the latency still decreased by 14\%. 

These results show that the pipeline of PersDet brings notable acceleration for the detector, making it suit for edge devices.

\subsection{Efficient BEV Representation}
\label{analy:eff}

\begin{table}[h]
	\begin{center}
		\scalebox{1}{
		    \begin{tabular}{c|ccc|c}
\toprule
Density & mAP & mATE & NDS & Mem. (MB)  \\
\midrule
$44\times56$ 
& 0.254 & 0.841 & 0.318 & 61\\
$88\times56$ 
& 0.281 & 0.768 & 0.336 & 83\\
$176\times56$ 
& \textbf{0.286} & 0.759 & 0.341 & 137\\
$352\times56$ 
& 0.277 & \textbf{0.758} & \textbf{0.342} & 247\\
\bottomrule
            \end{tabular}
        }
    \caption{Performance of CaDDN under different sample density.}
    \label{table:scale}
    \end{center}
\end{table}

According to Tab.~\ref{table:scale}, when the CaDDN detector achieves its best performance, it takes twice as much memory as PersDet needs. This colossal memory consumption is caused by inefficient representation. 
As shown in Fig.~\ref{fig:depth}, the regular BEV feature leaves a large portion of space as ``invalid''. The existence of these areas limited the space for valuable information. In that case, larger overall memory space is required for the same amount of information.
This observation indicates that the perspective BEV space enjoys the benefit of memory efficiency; thus, the PersDet achieves better performance while requiring less memory space.

\subsection{Ablation Study}

\begin{table}[h]
	\begin{center}
		\scalebox{1}{
		    \begin{tabular}{c|c|ccccccc}
\toprule
L.Y.       & Dir.Cls.     & mAP & mAOE & mAAE & NDS  \\
\midrule
$\times$        & $\times$     & 
\textbf{0.309} & 0.809 & 0.337 & 0.339 \\
$\times$        & $\checkmark$ & 
0.298 & 0.805 & \textbf{0.282} & 0.336 \\
$\checkmark$    & $\checkmark$ & 
0.300 & \textbf{0.678} & 0.312 & \textbf{0.348} \\
\bottomrule
            \end{tabular}
        }
    \caption{Performance boost of PersDet with our improved learning target. L.Y. denotes replace yaw angel with local yaw. Dir.Cls. denotes using direction classification.}
    \label{table:head}
    \end{center}
\end{table}

\subsubsection{Local Yaw and Categorical Direction}
In PersHead, we use local yaw and categorical direction derived from FCOS3D~\cite{fcos3d} to predict the yaw angle of objects. 
We show the different performances of the head under several settings in Tab.~\ref{table:head}. As can be seen in the table, the categorical direction reduces the Average Attribute Error (mAAE). This indicates that categorical direction also helps the model to classify objects better. When changing the yaw angle target to local yaw, the Average Orientation Error (mAOE) is further improved since the local yaw exactly corresponds to the image pattern.

\begin{table}[h]
	\begin{center}
		\scalebox{1}{
		    \begin{tabular}{c|ccccc}
\toprule
 DepthBins & mAP & mATE & mAOE & mAAE & NDS  \\
\midrule
14 & 0.272 & 0.823 & 0.645 & \textbf{0.282} & 0.332 \\
 28  & \textbf{0.304} & 0.751 & \textbf{0.637} & 0.286 & \textbf{0.356} \\
 56 & 0.300 & 0.735 & 0.678 & 0.312 & 0.348 \\
 112 & 0.290 & \textbf{0.725} & 0.693 & 0.327 & 0.342 \\
\bottomrule
            \end{tabular}
        }
    \caption{Performances of PersDet under different depth division. Models are trained \emph{without} CBGS.}
    \label{table:depth}
    \end{center}
\end{table}

\subsubsection{Number of Depth Categories}
Since the partition density of depth affects the density of anchors, the granularity of the DepthNet, and the receptive field of a single BEV grid, we conduct experiments on this hyper-parameter. As the experimental results in Tab.~\ref{table:depth} show, the model achieves the best mAP and NDS when dividing the depth into 28 bins. However, the Transition Error (mATE) increased by 1.6\% compared with the 56 bins setting due to inaccurate depth prediction. 
When the partition becomes denser, a lower Transition Error is obtained at the cost of lower mAP. In that case, the division granularity of depth is a trade-off between two metrics.

\section{Conclusion}
In this paper, we analyze the BEV paradigm for 3D object detection and point out that the essential difference between BEV and Camera-View detectors is the distribution of features on the depth dimension. 
By revisiting the view-transformation procedure that generates BEV features, we propose PersDet, which performs object detection on perspective BEV features without feature sampling. We demonstrate that the PersDet solved problems caused by feature sampling in existing BEV detectors and achieved state-of-the-art performance among monocular 3D detectors. According to detailed results and ablation studies, our approach performs better than existing methods while being simpler and deploy-friendly for edge devices.

\clearpage
\bibliography{persdet}

\begin{thebibliography}{28}
\providecommand{\natexlab}[1]{#1}

\bibitem[{Caesar et~al.(2020)Caesar, Bankiti, Lang, Vora, Liong, Xu, Krishnan,
  Pan, Baldan, and Beijbom}]{nuscenes}
Caesar, H.; Bankiti, V.; Lang, A.~H.; Vora, S.; Liong, V.~E.; Xu, Q.; Krishnan,
  A.; Pan, Y.; Baldan, G.; and Beijbom, O. 2020.
\newblock nuscenes: A multimodal dataset for autonomous driving.
\newblock In \emph{Proceedings of the IEEE/CVF conference on computer vision
  and pattern recognition}, 11621--11631.

\bibitem[{Chen et~al.(2016)Chen, Kundu, Zhang, Ma, Fidler, and
  Urtasun}]{monodet}
Chen, X.; Kundu, K.; Zhang, Z.; Ma, H.; Fidler, S.; and Urtasun, R. 2016.
\newblock Monocular 3d object detection for autonomous driving.
\newblock In \emph{Proceedings of the IEEE conference on computer vision and
  pattern recognition}, 2147--2156.

\bibitem[{Geiger et~al.(2015)Geiger, Lenz, Stiller, and Urtasun}]{kitti}
Geiger, A.; Lenz, P.; Stiller, C.; and Urtasun, R. 2015.
\newblock The KITTI vision benchmark suite.
\newblock \emph{URL http://www. cvlibs. net/datasets/kitti}, 2.

\bibitem[{He et~al.(2016)He, Zhang, Ren, and Sun}]{resnet}
He, K.; Zhang, X.; Ren, S.; and Sun, J. 2016.
\newblock Deep residual learning for image recognition.
\newblock In \emph{Proceedings of the IEEE conference on computer vision and
  pattern recognition}, 770--778.

\bibitem[{Huang and Huang(2022)}]{bevdet4d}
Huang, J.; and Huang, G. 2022.
\newblock Bevdet4d: Exploit temporal cues in multi-camera 3d object detection.
\newblock \emph{arXiv preprint arXiv:2203.17054}.

\bibitem[{Huang et~al.(2021)Huang, Huang, Zhu, and Du}]{bevdet}
Huang, J.; Huang, G.; Zhu, Z.; and Du, D. 2021.
\newblock Bevdet: High-performance multi-camera 3d object detection in
  bird-eye-view.
\newblock \emph{arXiv preprint arXiv:2112.11790}.

\bibitem[{J{\"o}rgensen, Zach, and Kahl(2019)}]{ss3d}
J{\"o}rgensen, E.; Zach, C.; and Kahl, F. 2019.
\newblock Monocular 3d object detection and box fitting trained end-to-end
  using intersection-over-union loss.
\newblock \emph{arXiv preprint arXiv:1906.08070}.

\bibitem[{Ku et~al.(2018)Ku, Mozifian, Lee, Harakeh, and
  Waslander}]{ku2018joint}
Ku, J.; Mozifian, M.; Lee, J.; Harakeh, A.; and Waslander, S.~L. 2018.
\newblock Joint 3d proposal generation and object detection from view
  aggregation.
\newblock In \emph{2018 IEEE/RSJ International Conference on Intelligent Robots
  and Systems (IROS)}, 1--8. IEEE.

\bibitem[{Ku, Pon, and Waslander(2019)}]{monodet2}
Ku, J.; Pon, A.~D.; and Waslander, S.~L. 2019.
\newblock Monocular 3d object detection leveraging accurate proposals and shape
  reconstruction.
\newblock In \emph{Proceedings of the IEEE/CVF conference on computer vision
  and pattern recognition}, 11867--11876.

\bibitem[{Lang et~al.(2019)Lang, Vora, Caesar, Zhou, Yang, and
  Beijbom}]{pointpillars}
Lang, A.~H.; Vora, S.; Caesar, H.; Zhou, L.; Yang, J.; and Beijbom, O. 2019.
\newblock Pointpillars: Fast encoders for object detection from point clouds.
\newblock In \emph{Proceedings of the IEEE/CVF conference on computer vision
  and pattern recognition}, 12697--12705.

\bibitem[{Li et~al.(2022)Li, Ge, Yu, Yang, Wang, Shi, Sun, and Li}]{bevdepth}
Li, Y.; Ge, Z.; Yu, G.; Yang, J.; Wang, Z.; Shi, Y.; Sun, J.; and Li, Z. 2022.
\newblock BEVDepth: Acquisition of Reliable Depth for Multi-view 3D Object
  Detection.
\newblock \emph{arXiv preprint arXiv:2206.10092}.

\bibitem[{Mousavian et~al.(2017)Mousavian, Anguelov, Flynn, and
  Kosecka}]{3dbbox}
Mousavian, A.; Anguelov, D.; Flynn, J.; and Kosecka, J. 2017.
\newblock 3d bounding box estimation using deep learning and geometry.
\newblock In \emph{Proceedings of the IEEE conference on Computer Vision and
  Pattern Recognition}, 7074--7082.

\bibitem[{Pan et~al.(2020)Pan, Sun, Leung, Andonian, and Zhou}]{vpn}
Pan, B.; Sun, J.; Leung, H. Y.~T.; Andonian, A.; and Zhou, B. 2020.
\newblock Cross-view semantic segmentation for sensing surroundings.
\newblock \emph{IEEE Robotics and Automation Letters}, 5(3): 4867--4873.

\bibitem[{Paszke et~al.(2019)Paszke, Gross, Massa, Lerer, Bradbury, Chanan,
  Killeen, Lin, Gimelshein, Antiga et~al.}]{pytorch}
Paszke, A.; Gross, S.; Massa, F.; Lerer, A.; Bradbury, J.; Chanan, G.; Killeen,
  T.; Lin, Z.; Gimelshein, N.; Antiga, L.; et~al. 2019.
\newblock Pytorch: An imperative style, high-performance deep learning library.
\newblock \emph{Advances in neural information processing systems}, 32.

\bibitem[{Philion and Fidler(2020)}]{lss}
Philion, J.; and Fidler, S. 2020.
\newblock Lift, splat, shoot: Encoding images from arbitrary camera rigs by
  implicitly unprojecting to 3d.
\newblock In \emph{European Conference on Computer Vision}, 194--210. Springer.

\bibitem[{Reading et~al.(2021)Reading, Harakeh, Chae, and Waslander}]{caddn}
Reading, C.; Harakeh, A.; Chae, J.; and Waslander, S.~L. 2021.
\newblock Categorical depth distribution network for monocular 3d object
  detection.
\newblock In \emph{Proceedings of the IEEE/CVF Conference on Computer Vision
  and Pattern Recognition}, 8555--8564.

\bibitem[{Roddick and Cipolla(2020)}]{pon}
Roddick, T.; and Cipolla, R. 2020.
\newblock Predicting semantic map representations from images using pyramid
  occupancy networks.
\newblock In \emph{Proceedings of the IEEE/CVF Conference on Computer Vision
  and Pattern Recognition}, 11138--11147.

\bibitem[{Roddick, Kendall, and Cipolla(2018)}]{oft}
Roddick, T.; Kendall, A.; and Cipolla, R. 2018.
\newblock Orthographic feature transform for monocular 3d object detection.
\newblock \emph{arXiv preprint arXiv:1811.08188}.

\bibitem[{Shi et~al.(2020)Shi, Guo, Jiang, Wang, Shi, Wang, and Li}]{pvrcnn}
Shi, S.; Guo, C.; Jiang, L.; Wang, Z.; Shi, J.; Wang, X.; and Li, H. 2020.
\newblock Pv-rcnn: Point-voxel feature set abstraction for 3d object detection.
\newblock In \emph{Proceedings of the IEEE/CVF Conference on Computer Vision
  and Pattern Recognition}, 10529--10538.

\bibitem[{Shi, Wang, and Li(2019)}]{pointrcnn}
Shi, S.; Wang, X.; and Li, H. 2019.
\newblock Pointrcnn: 3d object proposal generation and detection from point
  cloud.
\newblock In \emph{Proceedings of the IEEE/CVF conference on computer vision
  and pattern recognition}, 770--779.

\bibitem[{Simonelli et~al.(2019)Simonelli, Bulo, Porzi, L{\'o}pez-Antequera,
  and Kontschieder}]{monoDis}
Simonelli, A.; Bulo, S.~R.; Porzi, L.; L{\'o}pez-Antequera, M.; and
  Kontschieder, P. 2019.
\newblock Disentangling monocular 3d object detection.
\newblock In \emph{Proceedings of the IEEE/CVF International Conference on
  Computer Vision}, 1991--1999.

\bibitem[{Tian et~al.(2019)Tian, Shen, Chen, and He}]{fcos}
Tian, Z.; Shen, C.; Chen, H.; and He, T. 2019.
\newblock Fcos: Fully convolutional one-stage object detection.
\newblock In \emph{Proceedings of the IEEE/CVF international conference on
  computer vision}, 9627--9636.

\bibitem[{Wang et~al.(2022)Wang, Xinge, Pang, and Lin}]{pgd}
Wang, T.; Xinge, Z.; Pang, J.; and Lin, D. 2022.
\newblock Probabilistic and geometric depth: Detecting objects in perspective.
\newblock In \emph{Conference on Robot Learning}, 1475--1485. PMLR.

\bibitem[{Wang et~al.(2021)Wang, Zhu, Pang, and Lin}]{fcos3d}
Wang, T.; Zhu, X.; Pang, J.; and Lin, D. 2021.
\newblock Fcos3d: Fully convolutional one-stage monocular 3d object detection.
\newblock In \emph{Proceedings of the IEEE/CVF International Conference on
  Computer Vision}, 913--922.

\bibitem[{Yan, Mao, and Li(2018)}]{second}
Yan, Y.; Mao, Y.; and Li, B. 2018.
\newblock Second: Sparsely embedded convolutional detection.
\newblock \emph{Sensors}, 18(10): 3337.

\bibitem[{Yin, Zhou, and Krahenbuhl(2021)}]{centerpoint}
Yin, T.; Zhou, X.; and Krahenbuhl, P. 2021.
\newblock Center-based 3d object detection and tracking.
\newblock In \emph{Proceedings of the IEEE/CVF conference on computer vision
  and pattern recognition}, 11784--11793.

\bibitem[{Zhou and Tuzel(2018)}]{voxelnet}
Zhou, Y.; and Tuzel, O. 2018.
\newblock Voxelnet: End-to-end learning for point cloud based 3d object
  detection.
\newblock In \emph{Proceedings of the IEEE conference on computer vision and
  pattern recognition}, 4490--4499.

\bibitem[{Zhu et~al.(2019)Zhu, Jiang, Zhou, Li, and Yu}]{cbgs}
Zhu, B.; Jiang, Z.; Zhou, X.; Li, Z.; and Yu, G. 2019.
\newblock Class-balanced grouping and sampling for point cloud 3d object
  detection.
\newblock \emph{arXiv preprint arXiv:1908.09492}.

\end{thebibliography}

\end{document}